\documentclass[sigplan,screen]{acmart}
\AtBeginDocument{%
  \providecommand\BibTeX{{%
    \normalfont B\kern-0.5em{\scshape i\kern-0.25em b}\kern-0.8em\TeX}}}
\renewcommand\footnotetextcopyrightpermission[1]{}
\usepackage{times}
\usepackage{soul}
\usepackage{url}
\usepackage{caption}
\usepackage{graphicx}
\usepackage{amsmath}
\usepackage{amsthm}

\usepackage{booktabs}
\usepackage{multirow}
\usepackage[switch]{lineno}

\usepackage[linesnumbered,lined,ruled]{algorithm2e}
\usepackage{subcaption}
\usepackage{paralist, tabularx}

\newtheorem{theorem}{Theorem}
\newtheorem{assumption}{Assumption}

\newtheorem{lemma}[theorem]{Lemma}

\begin{document}

\author{Zeke Xia}
\affiliation{%
  \institution{East China Normal University}
  \streetaddress{200062}
  \city{Shanghai}
  \country{China}
}

\author{Ming Hu}
\affiliation{%
  \institution{Nanyang Technological University}
  \city{Singapore}
  \country{Singapore}
}
\authornote{Corresponding authors}
\email{hu.ming.work@gmail.com}

\author{Dengke Yan}
\affiliation{%
  \institution{East China Normal University}
  \streetaddress{200062}
  \city{Shanghai}
  \country{China}
}

\author{Ruixuan Liu}
\affiliation{%
  \institution{East China Normal University}
  \streetaddress{200062}
  \city{Shanghai}
  \country{China}
}

\author{Anran Li}
\affiliation{%
  \institution{Nanyang Technological University}
  \city{Notre Dame}
  \country{USA}
}

\author{Xiaofei Xie}
\affiliation{%
  \institution{Singapore Management University}
  \city{Singapore}
  \country{Singapore}
}

\author{Mingsong Chen}
\affiliation{%
  \institution{East China Normal University}
  \streetaddress{200062}
  \city{Shanghai}
  \country{China}
}
\authornotemark[1]
\email{mschen@sei.ecnu.edu.cn}

\title{
KoReA-SFL: \textbf{\underline{K}}n\textbf{\underline{o}}wledge \textbf{\underline{Re}}play-based Split Federated Learning 
\textbf{\underline{A}}gainst Catastrophic Forgetting
}

\begin{abstract}

Although Split Federated Learning (SFL) is good at enabling knowledge sharing among resource-constrained clients, it suffers from the problem of low training accuracy due to the neglect of data heterogeneity and catastrophic forgetting.
To address this issue, we propose a novel SFL approach named KoReA-SFL, which adopts a multi-model aggregation mechanism to alleviate gradient divergence caused by heterogeneous data and a knowledge replay strategy to deal with catastrophic forgetting. Specifically, in  KoReA-SFL cloud servers (i.e., fed server and main server) maintain multiple branch model portions rather than a global portion for local training and an aggregated master-model portion for knowledge sharing among branch portions. To avoid catastrophic forgetting, the main server of KoReA-SFL selects multiple assistant devices for knowledge replay according to the training data distribution of each server-side branch-model portion.
Experimental results obtained from non-IID and IID scenarios demonstrate that KoReA-SFL significantly outperforms conventional SFL methods (by up to 23.25\%  test accuracy improvement).


\end{abstract}

\maketitle

\section{Introduction}

As a promising distributed machine learning paradigm, Federated Learning (FL)~\cite{fedavg,li2020federated,hu2024fedmut}
enables numerous clients to collaboratively train Deep Neural Network (DNN) models without sharing their own data. 
However, due to resource constraints of heterogeneous devices, FL has limited application in  Artificial Intelligence of Things (AIoT) scenarios~\cite{hu2023aiotml,zhang2020efficient}.
Typically, FL server has to initialize a small model architecture that can be successfully deployed on FL clients with the fewest resources. 
Training with small FL models will inevitably result in numerous prediction errors and fail to fully utilize the resources of well-resourced clients. 
To facilitate the training of a high-performance model on heterogeneous devices, Split Federated Learning (SFL)~\cite{sfl} has been proposed. 
SFL splits the model into two portions, i.e., the client-side portion and the server-side portion, and it adopts two cloud servers, i.e., fed-server and main-server, for the privacy considerations. 
During model training, each client inputs its raw data into the client-side portion for forward propagation and uploads the output features with labels to the main server.
The main server then performs forward/ backward propagation on the server-side portion and sends gradients to clients. Then, clients conduct backward propagation and upload client-side portions to fed-server for aggregation. 


Although SFL works well in resource-constrained scenarios, it still suffers from the problem of data heterogeneity and catastrophic forgetting~\cite{cf}.
Specifically, since the data among AIoT devices are non-Independent and Identically Distributed (non-IID)~\cite{sattler_tnnls_2020}, SFL suffers from the ``gradient divergence'' problem~\cite{karimireddy2020scaffold}, which results in performance degradation of the aggregated global model.
Moreover, since the server only selects partial clients for model training, this would exacerbate the problem of model forgetting, i.e., the model tends to forget what it has learned from previous training data. 
Existing SFL methods, however, cannot deal with these issues.
In general, existing FL approaches attempt to use knowledge distillation technologies~\cite{zhu2021data,lin2020ensemble}, client selection strategies~\cite{chen2020fedcluster}, global control variable~\cite{li2020federated,karimireddy2020scaffold}, or multi-model paradigm~\cite{fedcross,gitfl} to alleviate performance degradation caused by non-IID data.
Besides, they adopt model correction~\cite{luo2023gradma} and data augmentation strategy~\cite{xu2022acceleration} to solve the catastrophic forgetting problem.
However, since the majority of these methods require the cloud server or clients access to the complete global model, existing methods cannot be directly applied in SFL. 
Therefore, \textit{how to design a novel training strategy to deal with accuracy degradation caused by non-IID data and catastrophic forgetting has become a serious challenge in SFL.}



\begin{figure}[t] 
	\begin{center} 
		\includegraphics[width=0.45\textwidth]{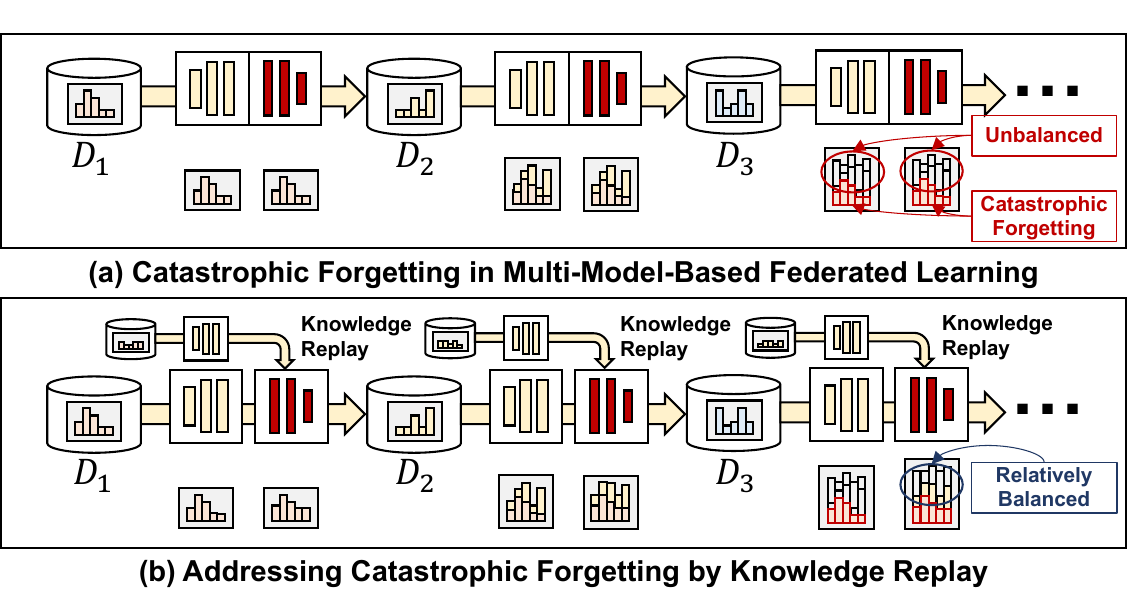}
 \vspace{-0.15in}
		\caption{Motivation of our approach.}
		\label{fig:motivation} 
	\end{center}
  \vspace{-0.3in}
\end{figure}

As a state-of-the-art paradigm, multi-model-based FL~\cite{fedcross,gitfl} adopts multiple homogeneous branch models for local training rather than the single global model for local training.
Specifically, each branch model selects the least similar model as the collaborative model for aggregation. To alleviate the gradient divergence, the collaborative model is assigned a relatively low weight for aggregation.
In addition, for each branch model, despite the fact that the distribution of data from each device is non-IID, the overall data distribution gradually converges towards balance through continuous training.
Intuitively, inspired by multi-model-based FL, the main server and the fed server can maintain multiple portions and assign an index to match the corresponding server-side and client-side portions to achieve training of multiple branch models.
Although the multi-model-based FL paradigm effectively alleviates the data heterogeneity problem, it still encounters the catastrophic forgetting problem.
Figure~\ref{fig:motivation}(a) presents an example of the training process of a branch model.
As shown in Figure~\ref{fig:motivation}(a), in non-IID scenarios, the branch model forgets the knowledge learned by $D_1$ after two training rounds and both $D_2$ and $D_3$ lack the data of the second category, which inevitably results in a degradation of the branch model classification performance for the second category.
Unlike conventional FL, although the main server cannot access the full model, it can access training data labels, which can be used to identify which categories of knowledge are missing for each branch model.
Intuitively, as shown in Figure~\ref{fig:motivation}(b), to alleviate catastrophic forgetting, the cloud server can request inactivated clients to upload a small number of missing category features to assist in training of the specific server-side portion.


Inspired by the above motivations, we propose a novel multi-model-based SFL approach named KoReA-SFL, which maintains multiple branch server-side and client-side portions for FL training.
To achieve knowledge sharing among multiple models, each server aggregates all the model portions to generate a master portion and aggregates the master portion with each branch portion, where the master portion is assigned a small aggregation weight.
To address catastrophic forgetting, the main server selects multiple inactivated clients as assistant clients and requests sampling features from these assistant clients for joint training for server-side portions according to the distribution of training data.
To improve training performance, KoReA-SFL adopts a sampling selection strategy to adaptively adjust the sampling proportion.
This paper has three major contributions:

\begin{itemize}
\item 
We present a novel multi-model-based  SFL framework named KoReA-SFL, which maintains a server-side portion repository and a client-side portion repository to enable multiple branch models to perform local training. By aggregating with an aggregated master model, branch models can share knowledge with each other.

\item 
We propose an adaptive knowledge-replay strategy that collects sampling features from inactivated clients according to the training data distribution of each branch model to alleviate catastrophic forgetting.
\item 
We conduct experiments on well-known datasets and models to show the effectiveness of our approach for both IID and non-IID scenarios.
\end{itemize}

\section{Related Work}
\label{sec:relatedwork}

{\bf Split Federated Learning}
Split Federated Learning~\cite{sfl} (SFL) combines the advantages of FL and Split Learning (SL). In SFL, the complete model is divided into two parts: the client-side model portion and the server-side model portion. Each client communicates directly with the main server and fed server. In each training round, clients interact with the main server in parallel to execute the SL process. Subsequently, clients send their updated client model portion to the fed server for aggregation. The fed server aggregates all clients' models and synchronizes the aggregated model with all clients. However, a shortcoming of SFL is that it still experiences the problem of low inference accuracy as in FL. The learning objective of SFL is to minimize the loss function over the collection of training data at $N$ clients, i.e.,
\begin{equation}
\footnotesize
    \min_{w} F(w)=\sum_{k=1}^{N}\frac{\left | D_k \right | }{\left |D  \right | } F_k(w), 
\end{equation}
where $w=w^c\oplus w^s$ is the combination of client-side model $w^c$ and server-side model $w^s$, $N$ is the number of clients that participate in local
training, $D_k$ is the dataset of the $k^{th}$ client, $F_k(w)=\frac{1}{\left | D_k \right | }  {\textstyle \sum_{j\in D_k}} f_j(w)$ is the loss empirical objective over the data samples at client $k$.

{\bf \noindent Catastrophic Forgetting}
Catastrophic forgetting occurs specifically when the neural network is trained sequentially on multiple tasks. In this case, the optimal parameters for the current task might perform poorly on the objectives of previous tasks. Many algorithms in FL have been proposed to alleviate the forgetting issue. GradMA~\cite{luo2023gradma} uses historical gradient information on both the device and server sides and corrects the global gradient through quadratic planning, effectively improving the accuracy of the global model. FedReg~\cite{xu2022acceleration} reduces knowledge forgetting by using generated pseudo data to regularize local training parameters and suppress potential conflicts with knowledge in local data introduced by pseudo data by using perturbed data. However, such methods cannot be applied to SFL due to the requirements of the complete model. To our best knowledge, KoReA-SFL represents the first approach that employs multi-model training and knowledge replay in SFL to enhance both model accuracy and training stability.

\section{Our KoReA-SFL Approach}
\label{sec:method}
\subsection{Overview}

\begin{figure*}[h] 
	\begin{center} 
		\includegraphics[width=0.95\textwidth]{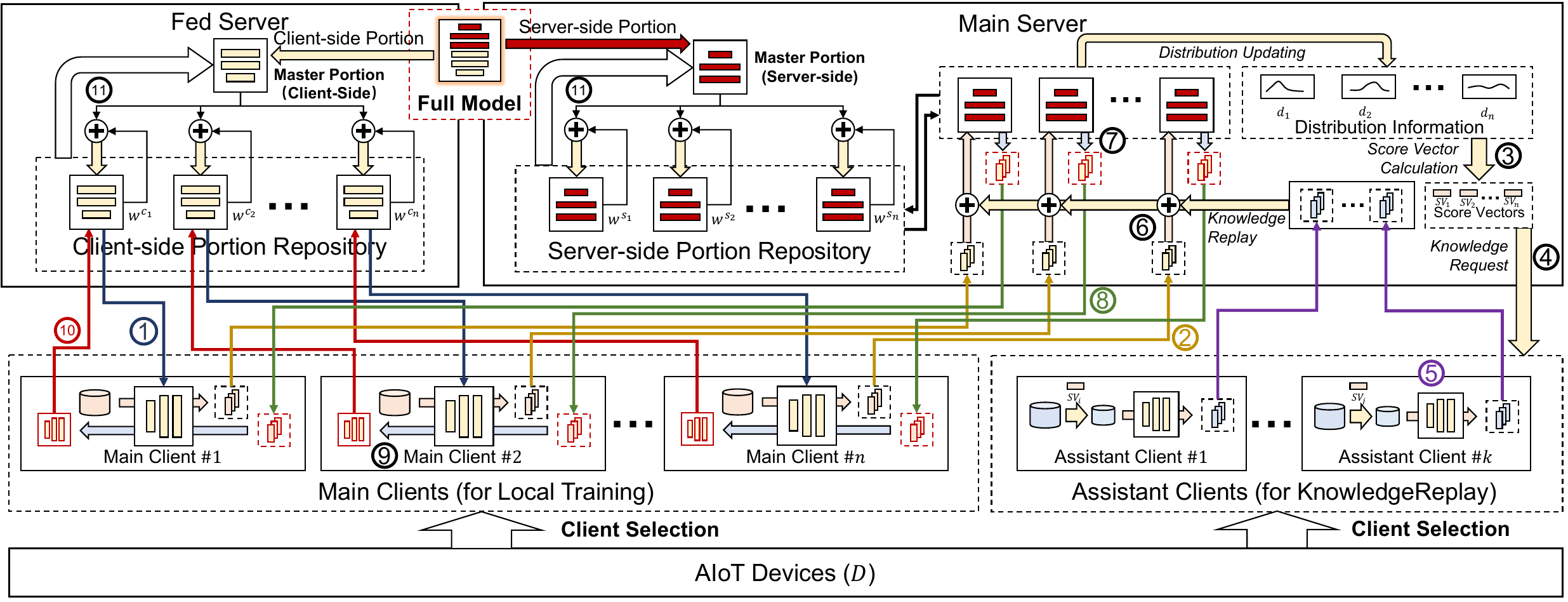}
 \vspace{-0.1in}
		\caption{Framework and workflow of our approach.}
		\label{fig:framework} 
 \vspace{-0.1 in}
	\end{center}
\end{figure*}

Figure \ref{fig:framework} illustrates the framework and workflow of our KoReA-SFL approach, which consists of two servers (i.e., the fed-server and the main-server) and multiple AIoT devices.
Similarly to conventional SFL, in KoReA-SFL, the full model is divided into two portions, i.e., the client-side portion and the server-side portion.
As shown in Figure \ref{fig:framework}, to address the gradient divergence problem caused by non-IID data, KoReA-SFL adopts multiple branch models for local training. 
To achieve multi-model-based FL training, the fed-sever maintains a client-side portion repository to store client-side branch portions and the main-server maintains a server-side portion repository to store server-side branch portions.
Note that a server-side and client-side portion corresponding to a branch model are assigned the same index in their repositories.
To enable knowledge sharing among branch portions, each server maintains a master portion, which is aggregated by all the portions in their repository.

To alleviate catastrophic forgetting, in each SFL training round, the main server calculates a score vector for each branch model, where the number of elements of the vector is equal to the number of classification categories. 
The value of each element is calculated according to the historical and current training data distribution, where the more recent training data distribution has a greater impact on the score vector calculation.
The main-server selects multiple inactivated clients as assistant clients and requests features of specific categories from assistant clients according to the score vectors to assist in the training of server-side portion.



As shown in Figure \ref{fig:framework}, the workflow of the training process for each intermediate model in KoReA-SFL includes eleven steps as follows:
\begin{compactitem}
    \item \textbf{Step 1 (Model Dispatching):} The fed server randomly selects $n$ clients as main clients to participate in local training and dispatches the branch client-side models to main clients for local training.
    \item \textbf{Step 2 (Feature Uploading): } Each main client uses its raw data to perform the forward propagation process and uploads the output features of its client-side portion to the main server.
    \item \textbf{Step 3 (Score Vector Calculation): } For each branch model, the main server calculates a score vector according to its current and historical training data distribution. 
    
    \item \textbf{Step 4 (Knowledge Request): } 
    Then the main server randomly selects $n$ inactivated clients as assistant clients, dispatches a calculated score vactor to an assistant client, and requests the fed server to dispatch client-portions to the corresponding assistant clients.
    \item \textbf{Step 5 (Knowledge Extraction): }
    Assistant clients select samples from their raw data according to the received score vector and use the selected samples to perform forward propagation.
    Then, they upload the output features to the main server.
    Note that if the main server does not collect enough features, it will repeat Step 3 to request more knowledge.
    
    \item \textbf{Step 6 (Knowledge Replay): }The main server inputs the features collected from the main and assistant clients together into the corresponding server-side portions for the forward propagation process and calculates the losses.

    \item \textbf{Step 7 (Server Model Backward): } The main server performs the backward propagation process to update the corresponding server-side portions and obtains the gradient of the features uploaded by the main clients.
    \item \textbf{Step 8 (Gradients Dispatching): } The main server sends the gradient to the corresponding main client.
    \item \textbf{Step 9 (Client Model Backward): } Each main client updates the client portion with the received gradient.
    \item \textbf{Step 10 (Model Uploading): } Each main client uploads its client-side portion to the fed server.
    \item \textbf{Step 11 (Model Repository Updating):} The fed server aggregates all client-side portions to generate a new master client-side portion and then updates each branch client-side portion by aggregating it with the master client-side portion. Similarly, the main server aggregates all the branch server-side portions to generate a new master server-side portion and then updates each branch server-side portion by aggregating it with the new master server-side portion. 
\end{compactitem}
\subsection{Implementation}

 Algorithm~\ref{alg:impl} presents the implementation of our KoReA-SFL approach. Assume that there are $n$ activated clients as main clients participating in local training at the same time. Lines 2-3 initialize the client-side model repository $W_c$ and its corresponding server-side model repository $W_s$. Lines 4-5 initialize model cumulative data distribution and sampling proportion. Lines 6-28 present the overall SFL training process. In line 7, we randomly select n clients as the main clients for local training. Lines 9-18 present the cooperative training process of each client-side model and its corresponding server-side model. In line 10, the main client $S[i]$ uses its local dataset to perform forward propagation and gain the intermediate feature $f_c$. Line 11 indicates that the main-server updates the data distribution of $w^{s_i}_r$. In line 12 the main-server calculates the score vector $sv^i$ based on the historical data distribution. In line 13, the main-server calculates the number of data to sample $q^i$ for each data class based on $sv^i$. Lines 14-20 present the process of sampling features. Firstly, the main-server initalizes the total sample supply $l^\prime$ the total sampled feature $f_h$ (Line 14). The main-server randomly selects an idle client $dev_a$ that can satisfy or partially satisfy the current knowledge request as an assistant client and the fed-server sends the client-side model $w^{c_i}_r$ to the $dev_a$ (Line 16). After receiving the $w^{c_i}_r$, $dev_a$ samples its local data and sends the feature of the sampled data using $w^{c_i}_r$ according to the current knowledge request $q^i-l^\prime$ (Line 17). After receiving the feature $f_a$, the main-server updates the total sample supply $l^\prime$ the total sampled feature $f_h$ (Lines 18-19). In line 21, the main-server puts $f_c$ and $f_s$ together. In lines 22-23, the main server uses $f_s$ to update the server-side model. Line 24 indicates that the main-server sends the gradient of $f_c$ back to the client $S[i]$, allowing $S[i]$ to update the client-side model $w_r^{c_i}$. Line 26 indicates that the fed-server adjusts the sample proportion for the next round. In lines 27-28, the fed-server and the main-server aggregate all client-side models and server-side models, respectively, to update $w_{r+1}^c$ and $w_{r+1}^s$. In lines 29-34, the fed-server and the main-server perform aggregation between the models in the model repositories and the global model and update the model repositories.

\begin{algorithm}[t]
    \caption{Our KoReA-SFL approach}
    \label{alg:impl} 
   \footnotesize
    \KwIn{
    i) $R$, maximum number of rounds; 
    ii) $C$, client set;
    }
    \KwOut{
    $w^g$, the global model
    }
    \textbf{KoReA-SFL}($R,C$) \textbf{begin}\;
    $W_c$$\leftarrow$[$w^{c_1}_0,w^{c_2}_0,...,w^{c_n}_0$]\;
    $W_s$$\leftarrow$[$w^{s_1}_0,w^{s_2}_0,...,w^{s_n}_0$]\;
    $L\leftarrow$[$L_1$,$L_2$,...,$L_n$]\;
    Initial $p_0\leftarrow0.01$\;
    \For{r=0,1,..,R-1}{
        $S\leftarrow$Randomly select $n$ devices from $C$\;
        /* parallel for */ \;
        \For{i=1,2,...,n}{
            $f_c,l_{S[i]}\leftarrow w^{c_i}_{r}(D_{S[i]})$\;
            $L_i[r] \leftarrow l_{S[i]}$\;
            $sv^i\leftarrow$SVCaculate($L_i$)\;
            $q^i\leftarrow$KnowledgeRequest($sv^i,p_r$)\;
            $l^\prime\leftarrow Zero(q^i)$;
            $f_h\leftarrow \{\}$\;
            \While{$l^\prime < q^i$}{
                $dev_a\leftarrow$Randomly select a device from $C-S$\; $f_a$,$l_a\leftarrow$KnowledgeExtract($w^{c_i}_{r}$,$D_{dev_a}$,$q^i-l^\prime$)\;
                $l^\prime\leftarrow l^\prime + l_a$\;
                $f_h\leftarrow f_h \oplus f_a$\;
            }
            $f_{s}\leftarrow f_c\oplus f_h$\;
            $y=w^{s_i}_{r}(f_s)$\;
            $w^{s_i}_{r}\leftarrow w^{s_i}_{r}-\eta\nabla (\ell  (y))$\;
            $w^{c_i}_{r}\leftarrow w^{c_i}_{r}-\eta\nabla (\frac{\partial \ell(y)}{\partial f_c} )$\;
        }
        $p_{r+1}=\frac{FGN(r)-FGN(r-1)}{FGN(r-1)} \times  p_r$\;
        $w^c_{r+1}\leftarrow  {\textstyle \sum_{i=1}^{n}\frac{w_{r+1}^{c_i}}{n}} $\;
        $w^s_{r+1}\leftarrow  {\textstyle \sum_{i=1}^{n}\frac{w_{r+1}^{s_i}}{n}} $\;
        \For{i=1,2,...,n}{
            $w^{c_i}_{r+1}\leftarrow (w_{r+1}^{c_i}+\alpha w_{r+1}^{c})/(1+\alpha)$\;
            $w^{s_i}_{r+1}\leftarrow (w_{r+1}^{s_i}+\alpha w_{r+1}^{s})/(1+\alpha)$\;
            $W_c[i]\leftarrow w^{c_i}_{r+1}$\;
            $W_s[i]\leftarrow w^{s_i}_{r+1}$\;
        }
    }
    $w^g \leftarrow w^c_{R}\oplus w^s_{R}$\;
    \textbf{Return} $w^g$\;
\end{algorithm}

\subsubsection{Knowledge Replay Strategy}


To solve catastrophic forgetting, KoReA-SFL adopts a knowledge replay strategy, which requests inactivated clients to upload features of categories that are less distributed in recent training data for each branch model and uses these features to train the corresponding server-side portions.
As shown in Algorithm~\ref{alg:impl}, our knowledge replay strategy consists of three key processes, i.e., score value calculation, knowledge request, and knowledge extraction.


\textbf{Score Vector Calculation (SVCalculate($\cdot$))} aims to evaluate the knowledge of each category learned by a target branch model, where the number of elements in a score vector equals to the number of categories and the model trained by more data of a specific category is assigned a larger score value for the corresponding category.
Based on the consideration of catastrophic forgetting, the current training data is assigned a larger weight for the score value calculation, while the historical training data are assigned lower weights.
For each branch model $w^i_r$, we calculate the score value $sv^i$ of $w^i_r$ as follows:
\begin{equation}
\footnotesize
    sv^i =\frac{ \sum_{j=0}^{r} \beta^{r-j} L_i[j] }{\sum_{j=0}^{r} \beta^{r-j}},
\end{equation}
where $\beta$ is the decay factor , and $l_j$ is the local data distribution of $w^i_r$ in the $r$-th round of training.

\textbf{Knowledge Request (KnowledgeRequest($\cdot$))} aims to calculate the number of data samples of each category that need to be selected to replay knowledge according to the score vector.
We first calculate the average score value of the score vector. The main server prefers to select the data of a category when its corresponding score value is larger than the average score.
We set a priority value for each data category.
For a data category $c$, its priority value is calculated as follows:
\begin{equation}
\footnotesize
prior^i_c=\max(0,mean(sv^i)-sv^i[c]).
\end{equation}
Note that the less accumulated data of data class $c$, the higher its priority. 
According to the priority values, the number of sampling data for the data category $c$ is calculated as follows:
\begin{equation}
\footnotesize
q^i_c = \frac{|D_i|\times p_r\times prior^i_c}{\sum_c prior^i_c}, 
\end{equation}
where $p_r$ is the sampling proportion of round $r$ and $|D_i|$ is the size of the local dataset of the client selected.

\textbf{Knowledge Extraction (KnowledgeExtract($\cdot$))}.
After receiving the $w^{c_i}_r$, the assistant client randomly selects its local data for forward propagation according to the calculated number of sampling data and uploads the intermediate features to the main-server. 
In addition, KoReA-SFL repeats the knowledge extraction process when the number of training data is lower than the number of sampling data.


    
\subsubsection{Adaptive Sample Proportion}

In our knowledge replay strategy, the sampling proportion $p_r$ is a key hyperparameter that affects the performance of KoReA-SFL.
Specifically, with increasing sampling proportion $p_r$, the main server requests more data for training the server-side portion, which can improve the accuracy of each branch model, especially in non-IID scenarios.
However, a larger value of $p_r$ inventively results in an increase in communication overhead.


To balance the accuracy of the model and communication overhead, we present a dynamic sampling proportion adjustment mechanism.
Previous work~\cite{CLP} observed that when the curvature of the loss landscape at a particular point $w$ is large, model training is in a critical learning period~\cite{CLP}.
Inspired by this observation, we prefer to select more data in the critical learning period and select fewer data at rather training rounds.
We use the Federated Gradient Norm (FGN) to approximate the curvature of the loss landscape at a particular point $w$ during training~\cite{criticalFL}. FGN of round $r$ can be defined as follows:
\begin{equation}
\footnotesize
    FGN(r)=  \frac{{\textstyle \sum_{i=1}^{n}-\eta \left \| g(w_r^i,\xi ) \right \| ^2}}{n},
\end{equation}
where $g(w_r^i,\xi )$ denotes the gradient of the loss function evaluated on $\xi$. 
Based on FGN, we adjust the sampling proportion $p_{t+1}$ for the next round as follows:
\begin{equation}
\footnotesize
    p_{r+1}=\frac{FGN(r)-FGN(r-1)}{FGN(r-1)} \times  p_r.
\end{equation}

\subsection{Convergence Analysis}
Our  analysis relies on the
following assumptions:

\begin{assumption}\label{asm1}
$f_i$ is $L$-smooth satisfying $f_i(w)\le f_i(w^\prime)+(w-w^\prime)^T\nabla f_i(w^\prime)+\frac{L}{2}||w-w^\prime||^2_2$, where $i \in\{ 1, 2, \cdots, N\}$.
\end{assumption}
\begin{assumption}\label{asm2}
$f_i$ is $\mu$-convex satisfying $f_i(w)\ge f_i(w^\prime)+(w-w^\prime)^T\nabla f_i(w^\prime)+\frac{\mu}{2}||w-w^\prime||^2_2$, where $i \in\{ 1, 2, \cdots, N\}$ and $\mu \geq 0$.
\end{assumption}
\begin{assumption}\label{asm3}

The variance of stochastic gradients is upper bounded by  $\sigma^2$ and the expectation of squared norm of stochastic gradients is upper bounded by  $G^2$, i.e.,  $\mathbb{E}||\nabla f_i (w;\xi) - \nabla f_i (w) ||^2 \leq \sigma^2$, $\mathbb{E}||\nabla f_i (w;\xi) ||^2 \leq G^2$, where $\xi$ is a  data batch of the $i^{th}$ client in the $t^{th}$ SFL round.
\end{assumption}

According to our aggregation strategy, we  have:
\begin{lemma}\label{eq:lemma1} Let $w_{r}^i= \alpha v_{r}^i + (1-\alpha)\overline{v}_{r}$, $\alpha\in [0,1]$, and $\overline{w}_r = \sum_{i=1}^N w_{r}^i$. We have
\begin{equation}
\footnotesize
\begin{split}
||\overline{w}_r - w^\star||^2\leq\frac{1}{N}\sum_{i=1}^N||w_{r}^i-w^\star||^2\leq\frac{1}{N}\sum_{i=1}^N||v_{r}^i-w^\star||^2,
\nonumber
\end{split}
\end{equation}
where $w^\star$ is the optimal parameters for the global loss function $F(\cdot)$. In other words, $\forall w, F^\star\leq F(w)$, where $F^\star$ denotes $F(w^\star)$. 
\end{lemma}

The following proves the convergence of our approach, whose details are 
available in  Appendix~\ref{sec:analysis}.

\newtheorem{thm}{\bf Theorem}
\begin{thm}\label{thm1}
Assume that the server performs model aggregation after $E$ rounds of SGD,i.e., $E$ rounds of SGD are performed in each SFL round., and the whole training consists of $r$ SFL rounds. Let $t=r\times  E$ be the current number of SGD rounds, and $\eta_t=\frac{2} {\mu (t + \lambda)}$ be the learning rate. We have: 
\begin{equation}
\label{eq:thm1}
\footnotesize
\begin{split}
\footnotesize
\mathbb{E}[F(\overline{w}_t)] -F^\star \leq \frac{L}{2\mu(t+\lambda)}\left[\frac{4B}{\mu} + \frac{\mu(\lambda+1)}{2}\Delta_1\right]
\end{split},
\end{equation}
where
\begin{small}
$
    B = 10 L \Gamma + 4(E - 1)^2 G^2.
     \nonumber
$
\end{small}
\end{thm}

\section{Experimental Results}
\label{sec:exp}
To evaluate
the effectiveness of our approach, we implemented KoReA-SFL using the PyTorch framework~\cite{pytorch} and conducted a comparative analysis with classical FL, i.e., FedAvg~\cite{fedavg} and SFL~\cite{sfl}. To ensure a fair comparison, we used an SGD optimizer with a fixed learning rate of 0.01 and a momentum of 0.5 for all baselines and KoReA-SFL, where each client was trained with a batch size of 50. All experimental results were obtained from an Ubuntu workstation equipped with  Intel i9 CPU, 64GB of memory, and  NVIDIA RTX 4090 GPU.

\subsection{Experimental Settings}
We compared KoReA-SFL with all baselines on four well-known datasets, i.e.,  CIFAR-10, CIFAR-100~\cite{cifar10}, FEMNIST~\cite{femnist}, and TinyImageNet~\cite{tinyimagenet}. To mimic the non-IID distributions of device data for CIFAR-10, CIFAR-100, and TinyImageNet, we employed the Dirichlet distribution $Dir(\beta)$ \cite{measuring}, where smaller values of $\beta$ indicate greater data heterogeneity. Note that 
the dataset FEMNIST itself is naturally non-IID distributed.
To show the pervasiveness of our approach, we conducted experiments on four widely-used DNN models, i.e., MobileNetV2 \cite{mobilenetv2}, ResNet-18 \cite{he2016deep}, VGG-16 \cite{vgg} and DenseNet-161~\cite{densenet}, respectively, which have different structures and depths. We simulated 100 clients and assumed that only 10\% of devices (i.e., $n=10$) participated in local training at the same time.
\subsection{Performance Comparison}

{\bf Comparison of Accuracy.}
Table \ref{tab:acc_main} compares the test accuracy between KoReA-SFL and all the baselines on four datasets with different non-IID and IID settings using the four DNN models. From Table \ref{tab:acc_main}, we can observe that our method (i.e.,  KoReA-SFL) achieves the highest test accuracy in all the scenarios regardless of models, datasets, and data heterogeneity. For example, our approach can improve test accuracy by 23.71\% over SFL on the CIFAR-10 dataset with the MobileNetV2 model when $\beta=0.1$.  As expected, our method achieves the best robustness in non-IID cases. 
For each model-dataset combination, we can find that the test accuracy gap increases significantly when the value of $\beta$ decreases. In other words, since our approach is good at addressing the problem of catastrophic forgetting,   the effect of our approach is more obvious when dealing with more non-IID scenarios.


\begin{table}[h]
\centering
\label{tab:acc_main}
\resizebox{\linewidth}{!}{%
\begin{tabular}{|c|c|c|c|c|c|}
\hline
\multirow{2}{*}{Model} & \multirow{2}{*}{Dataset} & \multirow{2}{*}{\begin{tabular}[c]{@{}c@{}}Hetero.\\ Settings\end{tabular}} & \multicolumn{3}{c|}{Test Accuracy (\%)} \\ \cline{4-6} 
 &  &  & FL & SFL & KoReA-SFL \\ \hline\hline
\multirow{9}{*}{MobileNetV2} & \multirow{4}{*}{CIFAR-10} & $\beta=0.1$ & $41.20\pm3.34$ & $40.74\pm3.72$ & $\textbf{64.45}\pm\textbf{0.33}$ \\
 &  & $\beta=0.3$ & $58.58\pm2.20$ & $54.57\pm2.51$ & $\textbf{69.23}\pm\textbf{0.56}$ \\
 &  & $\beta=0.5$ & $59.41\pm0.83$ & $58.69\pm0.88$ & $\textbf{70.44}\pm\textbf{0.53}$ \\
 &  & IID & $64.75\pm0.11$ & $63.80\pm0.18$ & $\textbf{75.87}\pm\textbf{0.16}$ \\ \cline{2-6} 
 & \multirow{4}{*}{CIFAR-100} & $\beta=0.1$ & $31.38\pm0.97$ & $31.07\pm1.28$ & $\textbf{45.66}\pm\textbf{0.27}$ \\
 &  & $\beta=0.3$ & $39.63\pm0.60$ & $39.37\pm0.55$ & $\textbf{45.36}\pm\textbf{0.61}$ \\
 &  & $\beta=0.5$ & $38.85\pm0.30$ & $40.26\pm0.60$ & $\textbf{46.34}\pm\textbf{0.67}$ \\
 &  & IID & $42.17\pm0.11$ & $40.63\pm0.22$ & $\textbf{52.53}\pm\textbf{0.13}$ \\ \cline{2-6} 
 & FEMNIST & - & $81.34\pm0.40$ & $80.92\pm0.35$ & $\textbf{82.63}\pm\textbf{0.28}$ \\ \hline
\multirow{9}{*}{ResNet18} & \multirow{4}{*}{CIFAR-10} & $\beta=0.1$ & $48.01\pm2.73$ & $45.08\pm3.64$ & $\textbf{66.72}\pm\textbf{0.48}$ \\
 &  & $\beta=0.3$ & $59.99\pm0.37$ & $60.13\pm1.01$ & $\textbf{69.69}\pm\textbf{0.20}$ \\
 &  & $\beta=0.5$ & $62.70\pm0.33$ & $63.00\pm0.34$ & $\textbf{70.61}\pm\textbf{0.24}$ \\
 &  & IID & $64.79\pm0.14$ & $64.51\pm0.22$ & $\textbf{73.20}\pm\textbf{0.07}$ \\ \cline{2-6} 
 & \multirow{4}{*}{CIFAR-100} & $\beta=0.1$ & $35.23\pm0.37$ & $35.71\pm0.52$ & $\textbf{42.38}\pm\textbf{0.22}$ \\
 &  & $\beta=0.3$ & $41.24\pm0.19$ & $41.30\pm0.22$ & $\textbf{46.23}\pm\textbf{0.18}$ \\
 &  & $\beta=0.5$ & $42.82\pm0.19$ & $43.08\pm0.13$ & $\textbf{47.86}\pm\textbf{0.16}$ \\
 &  & IID & $43.01\pm0.19$ & $43.10\pm0.22$ & $\textbf{48.83}\pm\textbf{0.12}$ \\ \cline{2-6} 
 & FEMNIST & - & $83.01\pm0.33$ & $83.40\pm0.24$ & $\textbf{84.18}\pm\textbf{0.17}$ \\ \hline
\multirow{9}{*}{VGG16} & \multirow{4}{*}{CIFAR-10} & $\beta=0.1$ & $66.26\pm4.55$ & $65.86\pm2.87$ & $\textbf{78.33}\pm\textbf{0.68}$ \\
 &  & $\beta=0.3$ & $77.23\pm0.26$ & $78.83\pm0.13$ & $\textbf{82.03}\pm\textbf{0.08}$ \\
 &  & $\beta=0.5$ & $78.47\pm0.09$ & $80.35\pm0.12$ & $\textbf{82.94}\pm\textbf{0.09}$ \\
 &  & IID & $79.92\pm0.07$ & $81.90\pm0.06$ & $\textbf{84.12}\pm\textbf{0.08}$ \\ \cline{2-6} 
 & \multirow{4}{*}{CIFAR-100} & $\beta=0.1$ & $47.70\pm1.63$ & $50.41\pm0.50$ & $\textbf{51.14}\pm\textbf{0.30}$ \\
 &  & $\beta=0.3$ & $53.93\pm0.44$ & $56.95\pm0.18$ & $\textbf{58.15}\pm\textbf{0.32}$ \\
 &  & $\beta=0.5$ & $55.00\pm0.59$ & $57.52\pm0.62$ & $\textbf{59.23}\pm\textbf{0.19}$ \\
 &  & IID & $57.26\pm0.07$ & $60.10\pm0.14$ & $\textbf{64.05}\pm\textbf{0.18}$ \\ \cline{2-6} 
 & FEMNIST & - & $82.24\pm0.57$ & $84.35\pm0.29$ & $\textbf{84.86}\pm\textbf{0.25}$ \\ \hline
\multirow{4}{*}{DenseNet161} & \multirow{4}{*}{Tiny-ImageNet} & $\beta=0.01$ & $21.64\pm0.34$ & $21.59\pm0.47$ & $\textbf{34.86}\pm\textbf{0.21}$ \\
 &  & $\beta=0.05$ & $33.65\pm0.42$ & $33.64\pm0.33$ & $\textbf{39.32}\pm\textbf{0.21}$ \\
 &  & $\beta=0.1$ & $35.71\pm0.27$ & $36.17\pm0.26$ & $\textbf{40.52}\pm\textbf{0.15}$ \\
 &  & IID & $38.53\pm0.11$ & $39.61\pm0.18$ & $\textbf{43.00}\pm\textbf{0.11}$ \\ \hline
\end{tabular}%
}
\caption{Comparison of test accuracy.}
\label{tab:acc_main}
\end{table}

\begin{figure}[h]
    \centering
    \begin{subfigure}[b]{0.22\textwidth}
        \includegraphics[width=\textwidth]{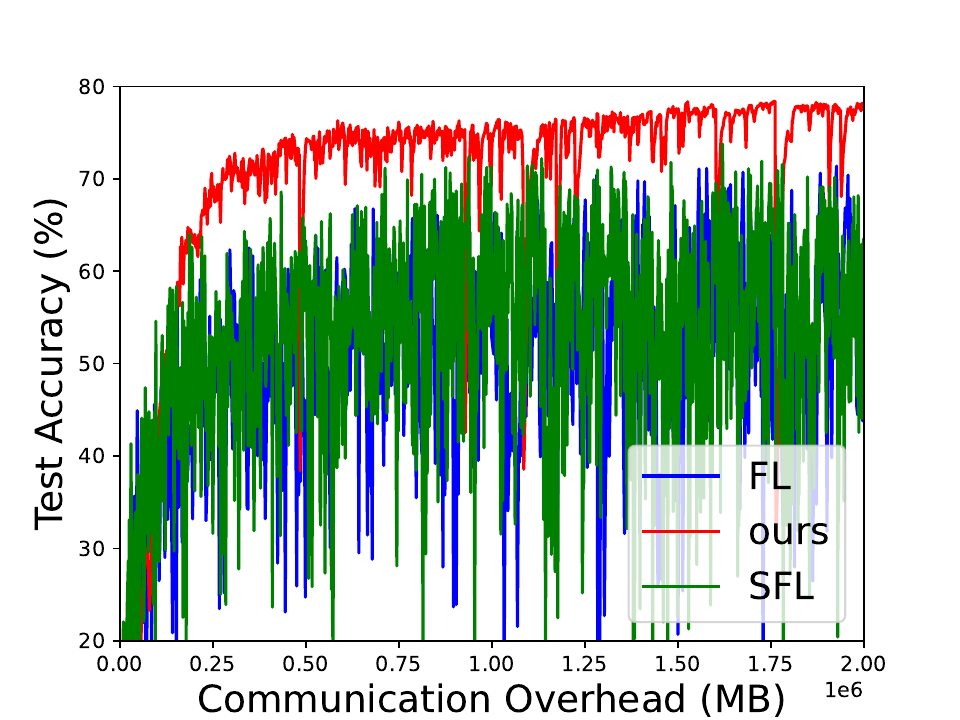}
        \caption{$\beta=0.1$}
        \label{fig:subfig1}
    \end{subfigure}
    \begin{subfigure}[b]{0.22\textwidth}
        \includegraphics[width=\textwidth]{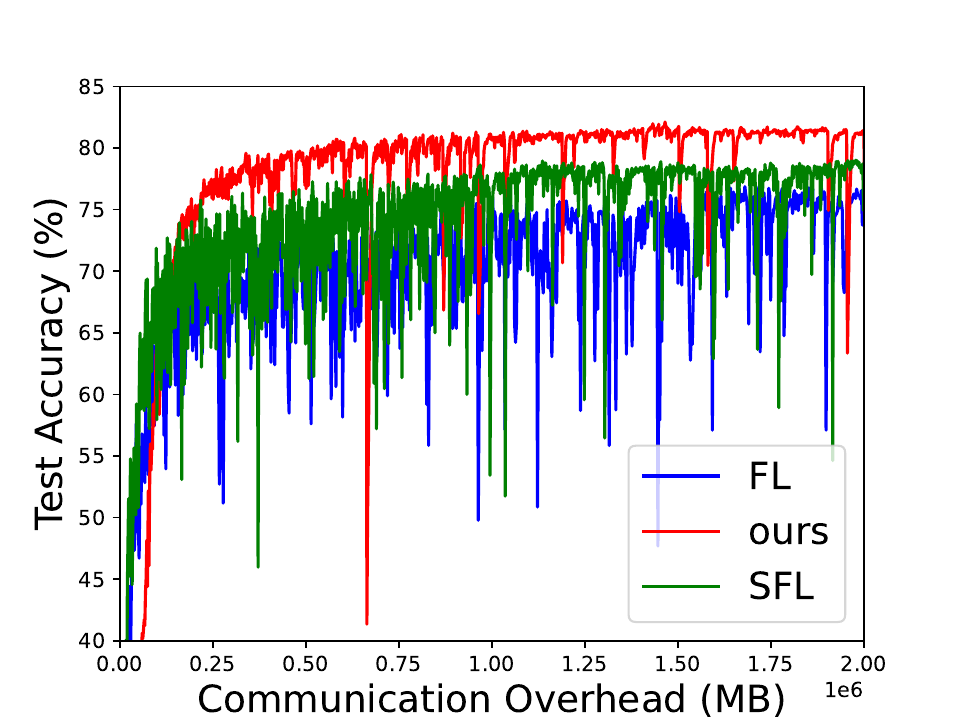}
        \caption{$\beta=0.3$}
        \label{fig:subfig2}
    \end{subfigure}
    \begin{subfigure}[b]{0.22\textwidth}
        \includegraphics[width=\textwidth]{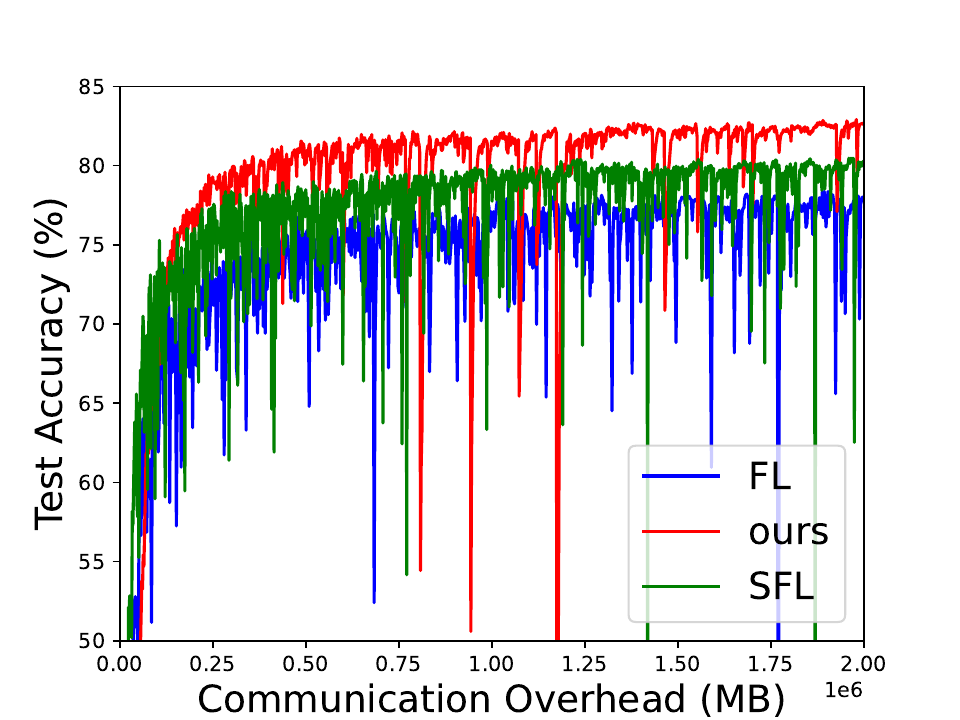}
        \caption{$\beta=0.5$}
        \label{fig:subfig3}
    \end{subfigure}
    \begin{subfigure}[b]{0.22\textwidth}
        \includegraphics[width=\textwidth]{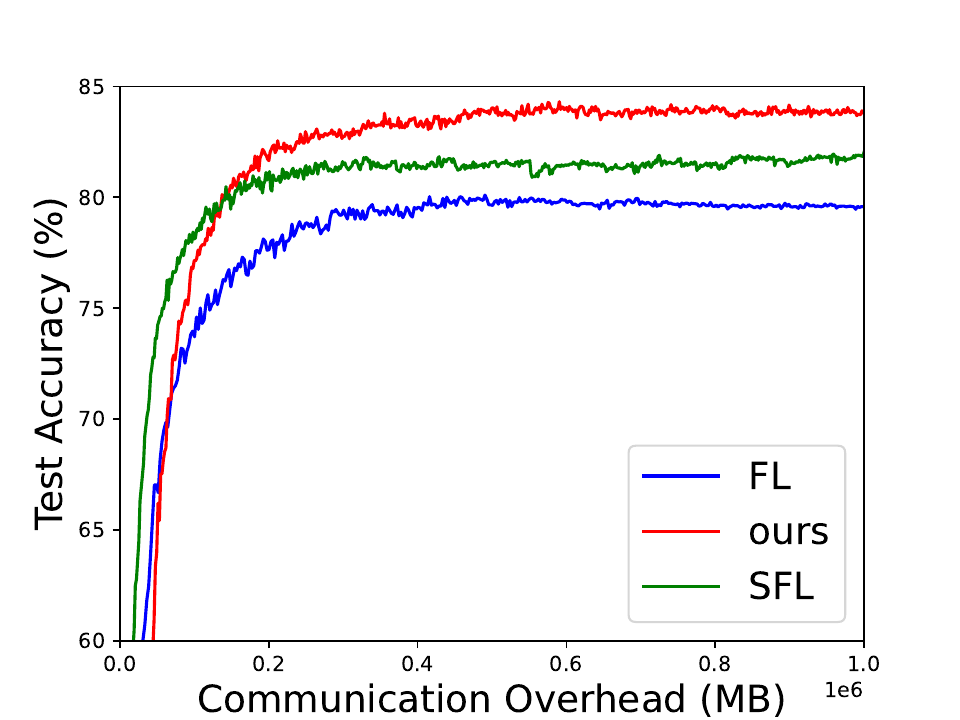}
        \caption{$IID$}
        \label{fig:subfig4}
    \end{subfigure}
    \caption{Comparison of communication overhead.}
    \label{fig:acc_compare}
\end{figure}

{\bf Comparison of Communication Overhead.}
Figure \ref{fig:acc_compare} presents the learning curves of our approach and all baseline methods on  CIFAR-10 using VGG-16, where the horizontal axes denote the communication overhead along the SFL training. We can observe that although KoReA-SFL introduces additional communication overhead caused by sampling features and assistant client models, our approach still achieves the highest accuracy for the same communication overhead. 
Meanwhile, we can observe that the learning curves of our method are much more stable than those of SFL and FL in scenarios with higher data heterogeneity.

\begin{figure}[th]
	\centering
      \begin{subfigure}[b]{0.23\textwidth}
        \includegraphics[width=\textwidth]{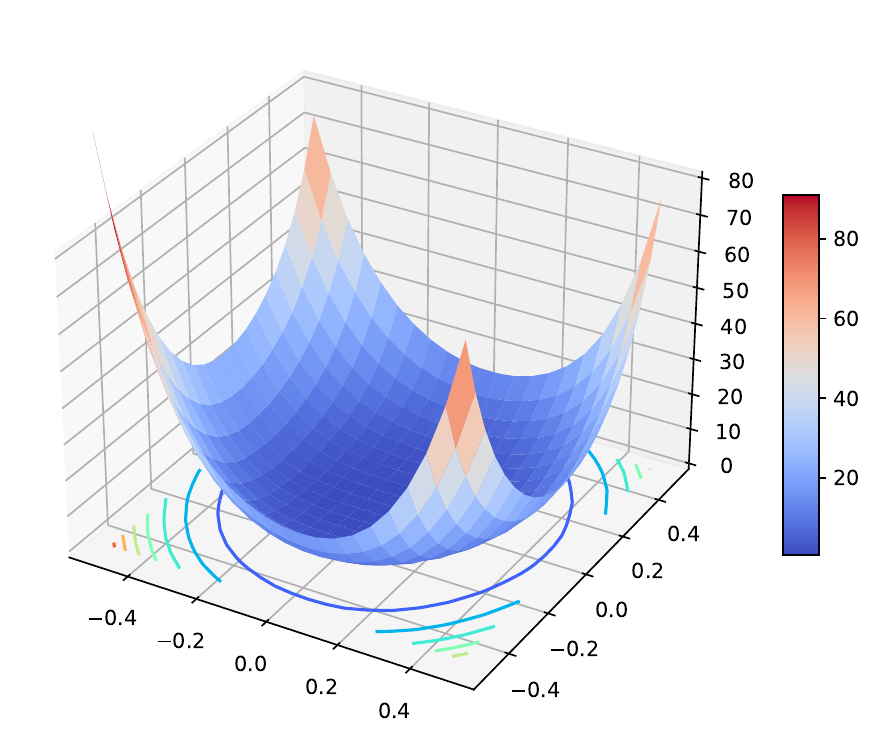}
        \caption{FL ($0.1$)}
        \label{fig:subfig1}
      \end{subfigure}
   \hspace{-0.15 in}
    \begin{subfigure}[b]{0.23\textwidth}
        \includegraphics[width=\textwidth]{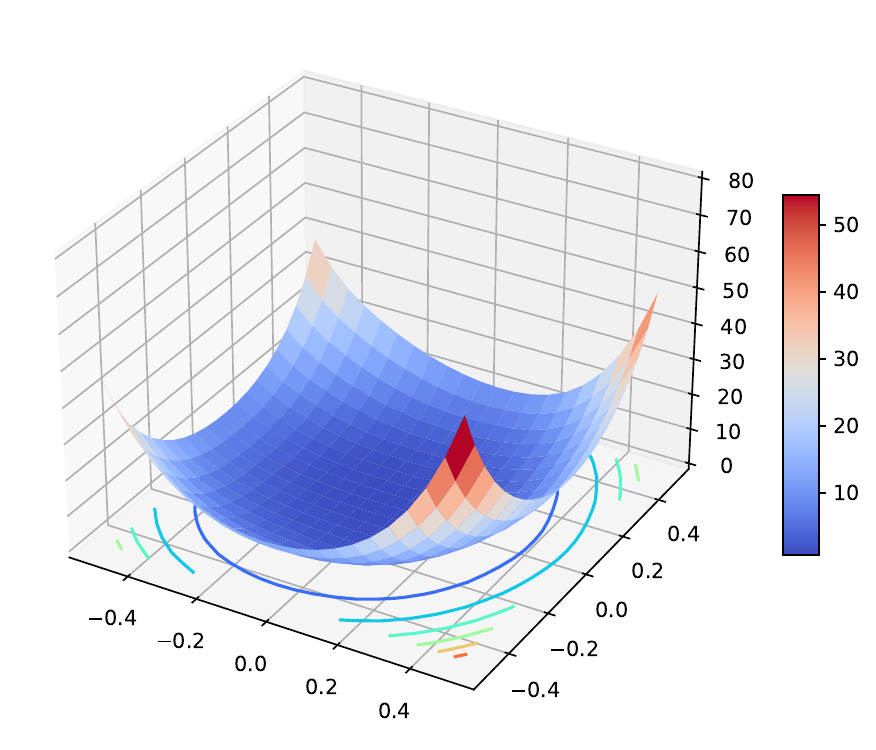}
        \caption{SFL ($0.1$)}
        \label{fig:subfig2}
      \end{subfigure}
      \begin{subfigure}[b]{0.23\textwidth}
        \includegraphics[width=\textwidth]{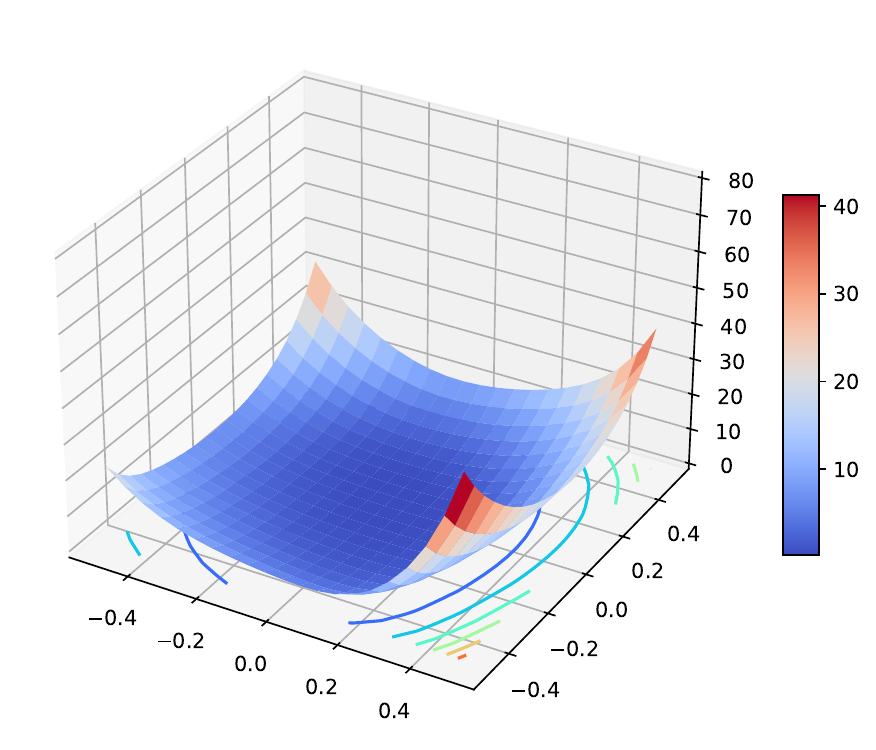}
        \caption{ours ($0.1$)}
        \label{fig:subfig3}
      \end{subfigure}
\hspace{-0.15 in}
      \begin{subfigure}[b]{0.23\textwidth}
        \includegraphics[width=\linewidth]{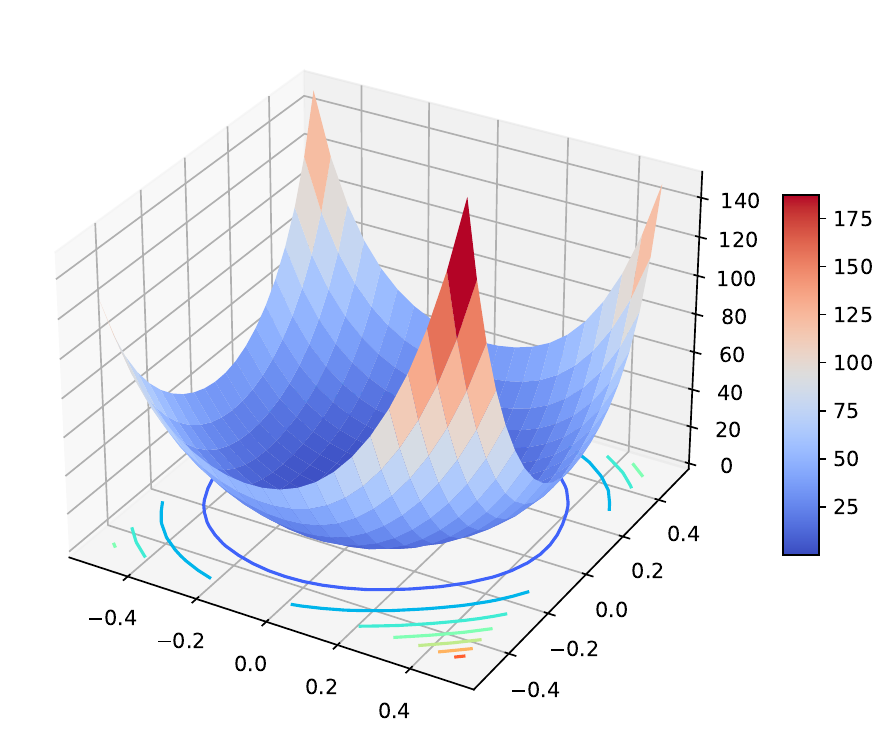}
        \caption{FL (IID)}\label{fig:subfig1}
      \end{subfigure}
      \begin{subfigure}[b]{0.23\textwidth}
        \includegraphics[width=\linewidth]{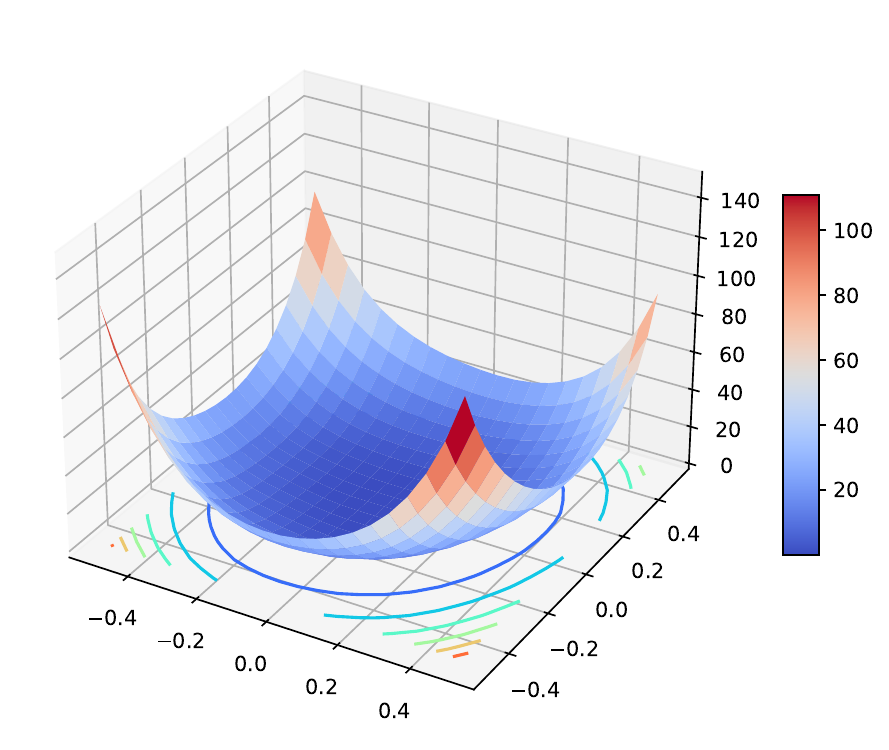}
        \caption{SFL (IID)}
        \label{fig:subfig2}
      \end{subfigure}
       \hspace{-0.15 in}
      \begin{subfigure}[b]{0.23\textwidth}
        \includegraphics[width=\linewidth]{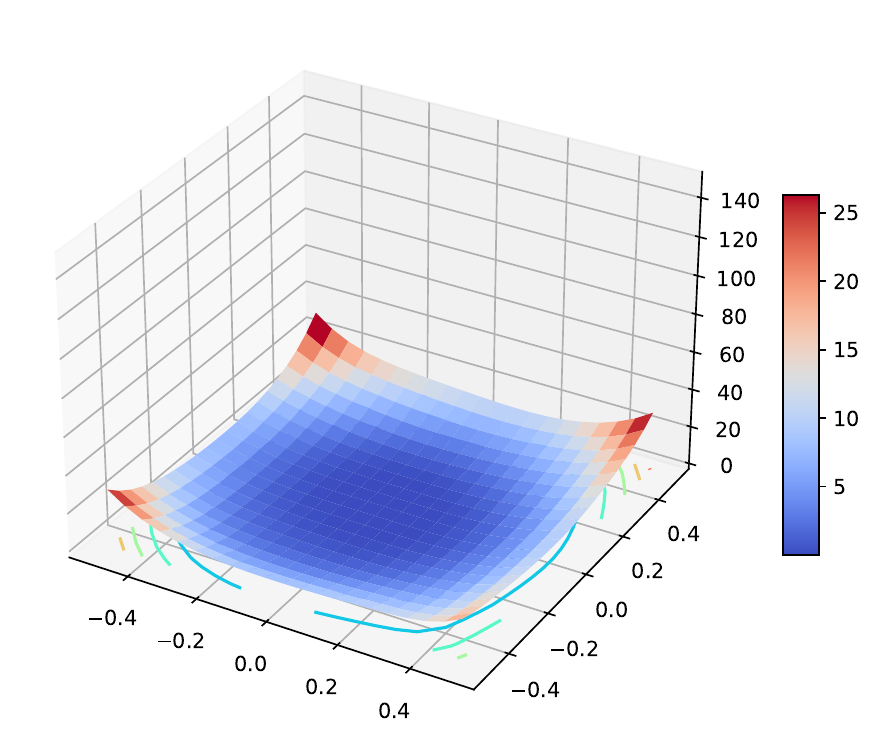}
        \caption{ours (IID)}
        \label{fig:subfig3}
      \end{subfigure}
	\caption{Comparison of loss landscapes.}
	\label{fig:loss}
\end{figure}

{\bf Comparison of Generalizability.}
To validate whether the global models trained by KoReA-SFL can converge into a flatter valley than the baselines, we plotted the loss landscapes of the three global models in 
Figure \ref{fig:loss} for ResNet-18 that are trained using FL, SFL, and our method on the CIFAR-10 dataset with $\beta=0.1$ and IID. 
We can find that, compared with FL and SFL, our method can achieve the lowest global losses, and the global models trained by our method are located in the flattest areas, indicating that the global models trained by our method achieve better generalization.

\subsection{Impacts of Different Configurations}
\begin{figure}[h]
    \centering
     \begin{subfigure}[b]{0.22\textwidth}
        \includegraphics[width=\textwidth]{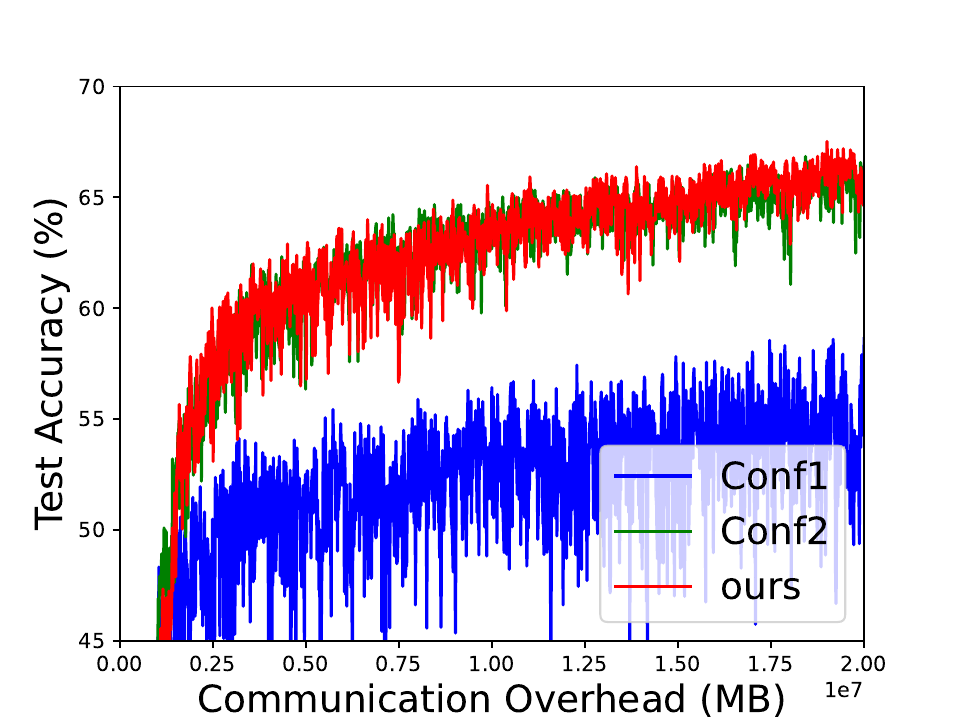}
        \caption{$\beta=0.1$}
        \label{fig:subfig1}
    \end{subfigure}
      \hspace{-0.1in}
    \begin{subfigure}[b]{0.22\textwidth}
        \includegraphics[width=\textwidth]{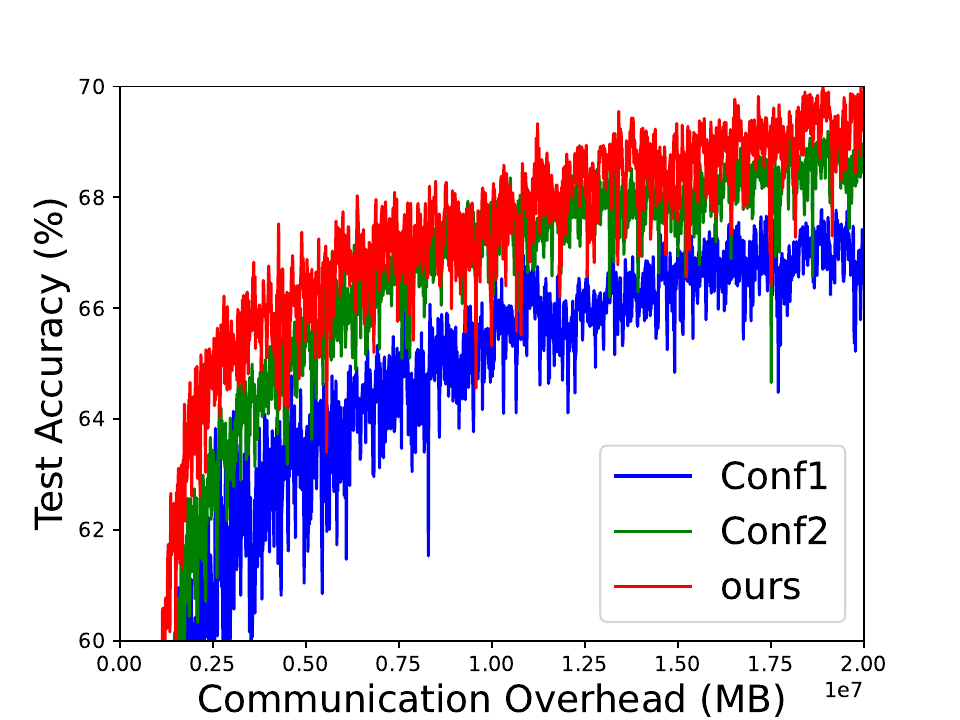}
        \caption{$\beta=0.3$}
        \label{fig:subfig2}
    \end{subfigure}
    
    \begin{subfigure}[b]{0.22\textwidth}
        \includegraphics[width=\textwidth]{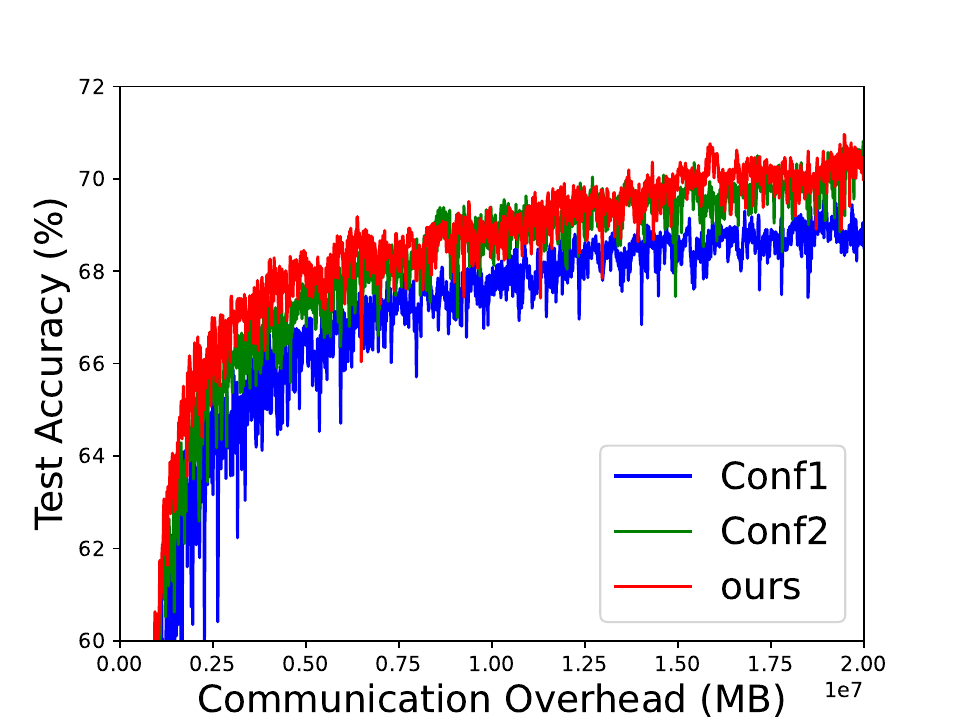}
        \caption{$\beta=0.5$}
        \label{fig:subfig3}
    \end{subfigure}
     \hspace{-0.1in}
    \begin{subfigure}[b]{0.22\textwidth}
        \includegraphics[width=\textwidth]{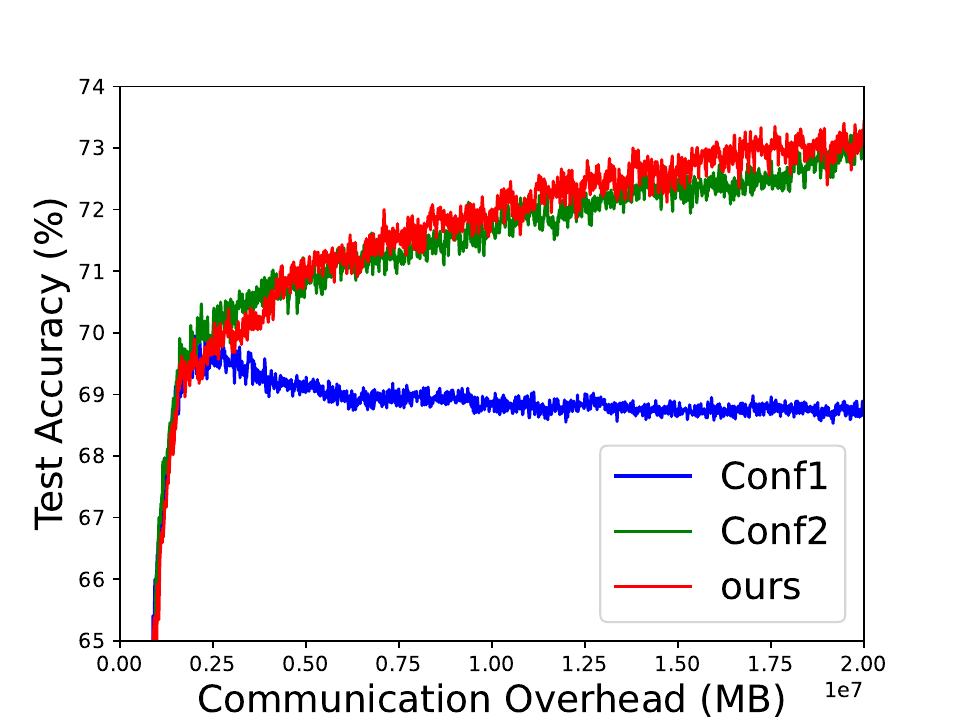}
        \caption{$IID$}
        \label{fig:subfig4}
    \end{subfigure}
    \caption{Learning curves for the different configurations.}
    \label{fig:fig_abliation}
\end{figure}

{\bf Ablation Study.}
To demonstrate the effectiveness of our proposed mechanisms in 
 KoReA-SFL, we developed two variants of KoReA-SFL: i) {\it Conf1} that represents our approach without knowledge replay; and ii) {\it Conf2} that denotes KoReA-SFL with knowledge replay and fixed feature sampling proportion. Note that {\it Conf2} has the same overall feature sampling proportion as KoReA-SFL, and {\it Conf1}, {\it Conf2}, and KoReA-SFL have the same total communication overhead. We conducted experiments using the ResNet-18 model on CIFAR-10 with both non-IID and IID settings, and the results are presented in Figure\ref{fig:fig_abliation} and Table \ref{abliation_table}.

 We can find that KoReA-SFL achieves the highest test accuracy among all three designs. From Figure \ref{fig:fig_abliation}, we can find that compared with 
 {\it Conf1}, both  {\it Conf1} and KoReA-SFL based on knowledge replay can achieve more stable learning curves with better test accuracy, especially when handling data distributions with higher heterogeneity. This again approves the effectiveness of our knowledge replay method. Compared with {\it Conf2}, we can find that,
 due to our proposed sampling proportion adaptive adjustment strategy, 
 the training processes of  KoReA-SFL are sped up to achieve slightly better inference performance.



 \begin{table}[h]
\centering
\scriptsize
\begin{tabular}{|c|c|c|c|}
\hline
\multirow{2}{*}{\begin{tabular}[c]{@{}c@{}}Hetero.\\ Settings\end{tabular}} & \multicolumn{3}{c|}{Test Accuracy (\%)} \\ \cline{2-4} 
& {Conf1} & Conf2   & {KoReA-SFL} \\ \hline\hline
$\beta=0.1$  & \multicolumn{1}{c|}{$56.00\pm1.27$} & $65.93\pm0.26$  & \multicolumn{1}{c|}{$\textbf{66.72}\pm\textbf{0.48}$}\\ \hline
$\beta=0.3$ &  \multicolumn{1}{c|}{$66.63\pm0.21$} & $68.15\pm0.24$  & \multicolumn{1}{c|}{$\textbf{68.80}\pm\textbf{0.20}$}\\ \hline
$\beta=0.5$ & \multicolumn{1}{c|}{$69.20\pm0.21$} & $70.50\pm0.21$ & \multicolumn{1}{c|}{$\textbf{70.61}\pm\textbf{0.24}$}  \\ \hline
IID  & \multicolumn{1}{c|}{$69.72\pm0.08$} & $73.10\pm0.11$ & \multicolumn{1}{c|}{$\textbf{73.20}\pm\textbf{0.07}$}\\ \hline
\end{tabular}
\caption{Ablation study results.}
\label{abliation_table}
\end{table}

\begin{figure}[h]
    \centering
    \begin{subfigure}[b]{0.22\textwidth}
        \includegraphics[width=\textwidth]{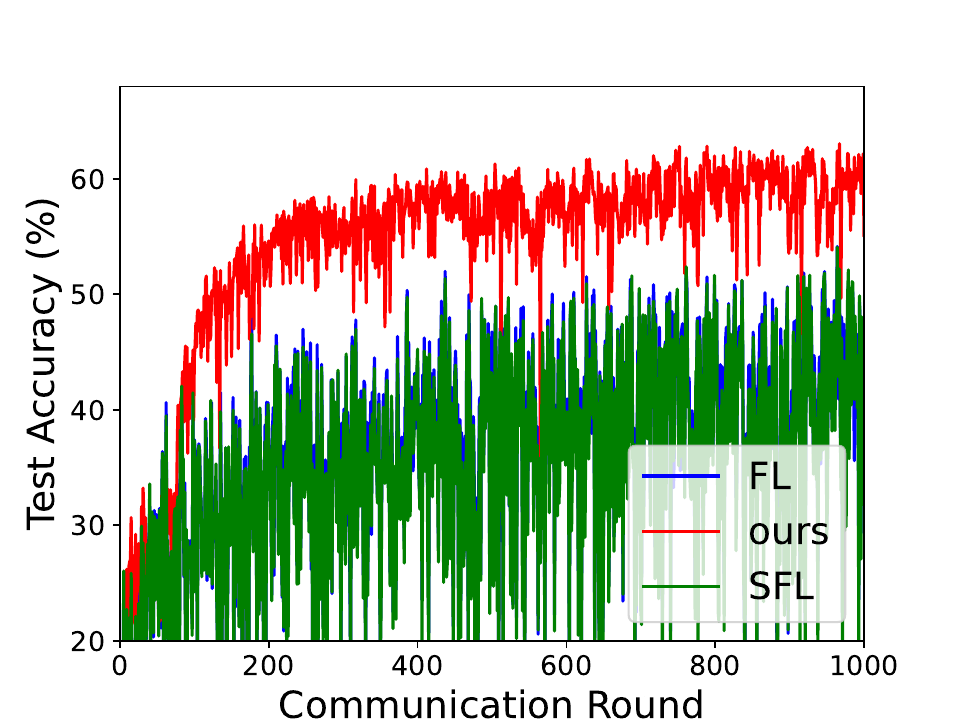}
        \caption{$n=5$}
    \end{subfigure}
      \hspace{-0.1in}
    \begin{subfigure}[b]{0.22\textwidth}
        \includegraphics[width=\textwidth]{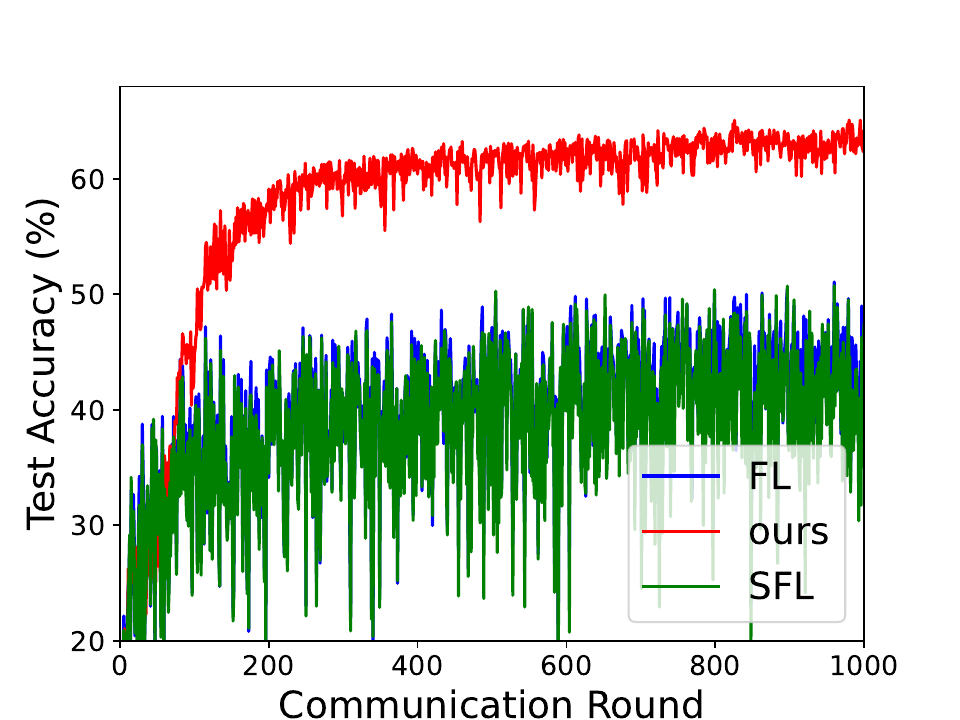}
        \caption{$n=10$}
    \end{subfigure}
    \hspace{-0.05in}
    \begin{subfigure}[b]{0.22\textwidth}
        \includegraphics[width=\textwidth]{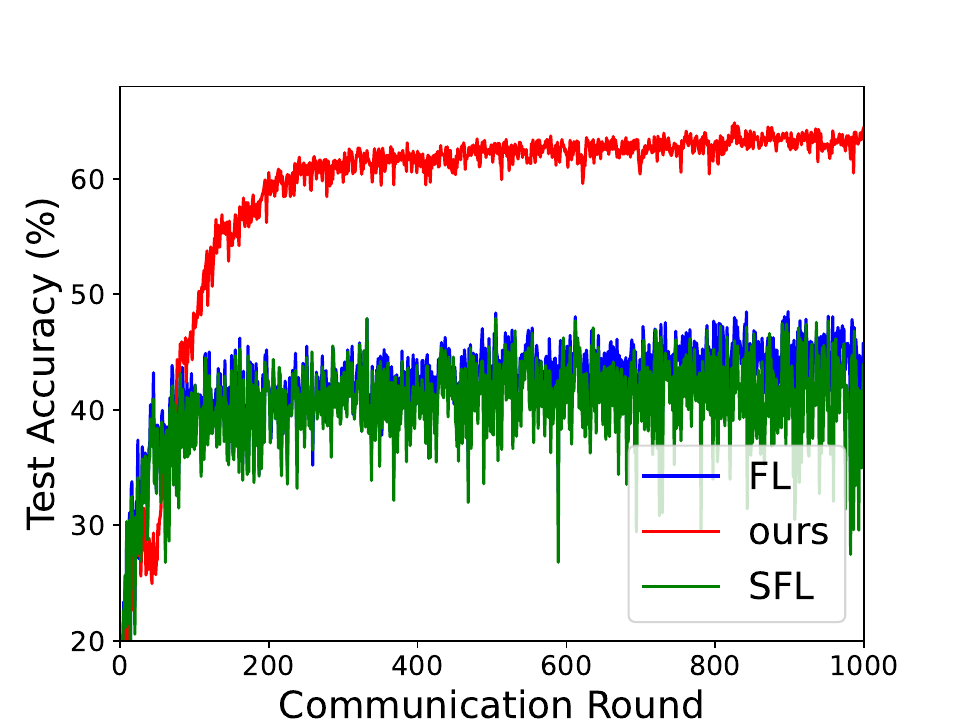}
        \caption{$n=20$}
    \end{subfigure}
      \hspace{-0.1in}
    \begin{subfigure}[b]{0.22\textwidth}
        \includegraphics[width=\textwidth]{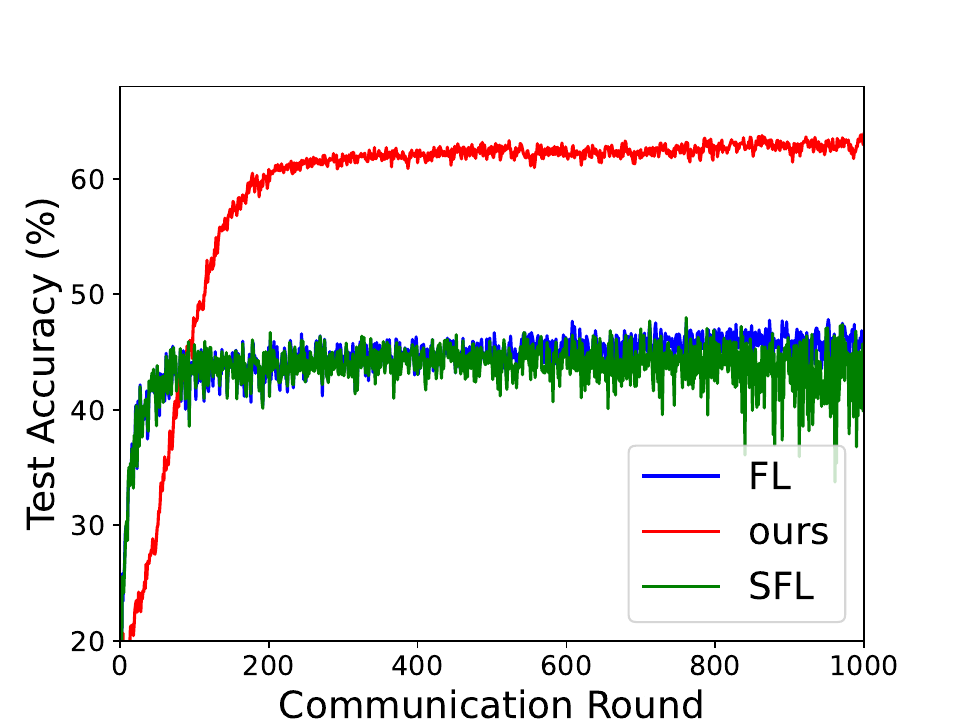}
        \caption{$n=50$}
    \end{subfigure}
    \caption{Learning curves considering different numbers of clients.}
    \label{fig:diff_k}
\end{figure}

{\bf Impacts of the Number of Participating Clients.}
In previous experiments, we assumed that 10\% of the devices participated in the SFL training at the same time. To investigate the impact of different numbers of devices trained simultaneously, we investigated
four different numbers (i.e., $n=5,10,20$ and $50$) of devices trained simultaneously
on CIFAR-10 using ResNet-18  with  $\beta=0.1$. Figure \ref{fig:diff_k} shows the experimental results. We can find that our approach can achieve the best test accuracy in all four cases while the learning curves of our method converge in the most stable manner, showing the superiority of our method in scalability.




\begin{figure}[h]
    \centering
    \begin{subfigure}[b]{0.22\textwidth}
        \includegraphics[width=\textwidth]{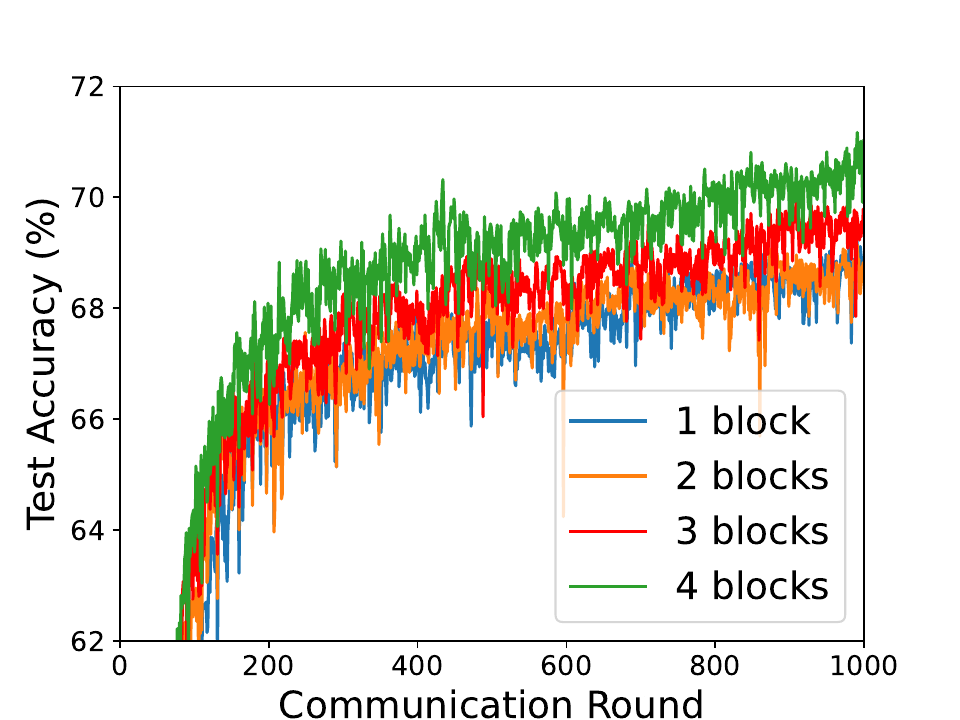}
        \caption{$\beta=0.5$}
    \end{subfigure}
    \begin{subfigure}[b]{0.22\textwidth}
        \includegraphics[width=\textwidth]{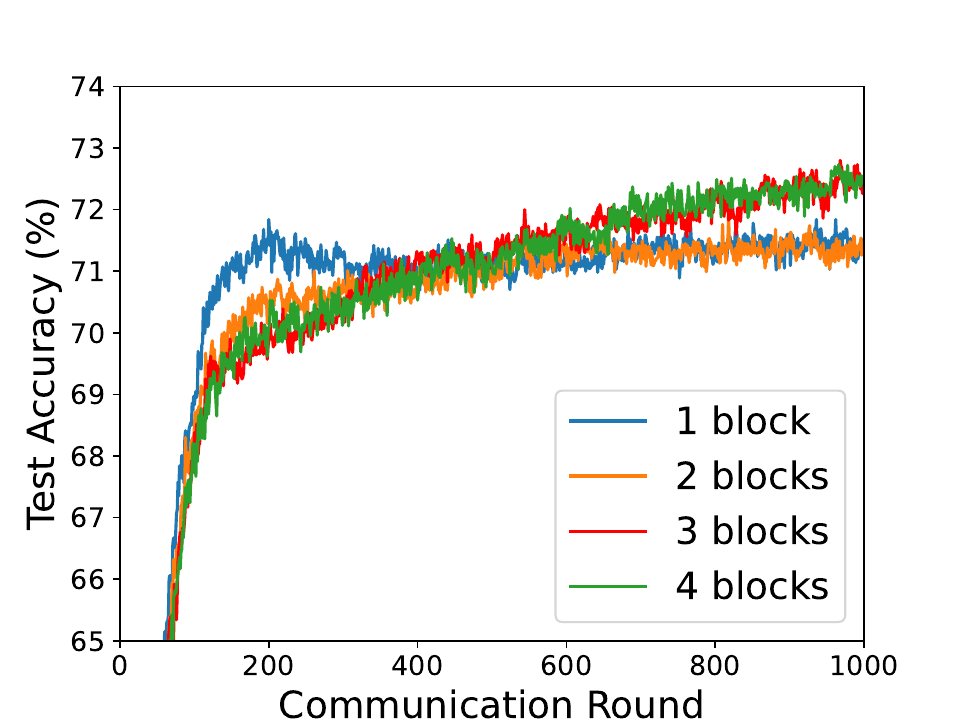}
        \caption{$IID$}
    \end{subfigure}
    \caption{Learning curves for cases w/ different splitting positions.}
    \label{fig:diff_sp}
\end{figure}

{\bf Impacts of Model Splitting Position.}
To evaluate the impacts of the server-side model share on the global model performance, we conducted an experiment on CIFAR-10 using Resnet-18, considering both IID and non-IID ($\beta=0.5$) scenarios.
Since ResNet-18 has four blocks, we investigated four different splitting positions, i.e., 1-4 blocks are kept in the server-side models, respectively. 
Figure \ref{fig:diff_sp} shows the experimental results. We can observe that the larger the proportion of server-side models, the better accuracy the global model can achieve, since our approach focuses on improving the performance of server-side models. Note that, for SFL, a larger proportion of server-side models may lead to larger transmission feature sizes, thus inevitably
resulting in higher communication overhead. 
Conversely, a larger proportion of client-side models may lead to less communication overhead, but the training performance/convergence may be greatly affected
due to the limited computing power of client devices. 


\begin{figure}[h]
  \begin{center} 
		\includegraphics[width=0.9\linewidth]{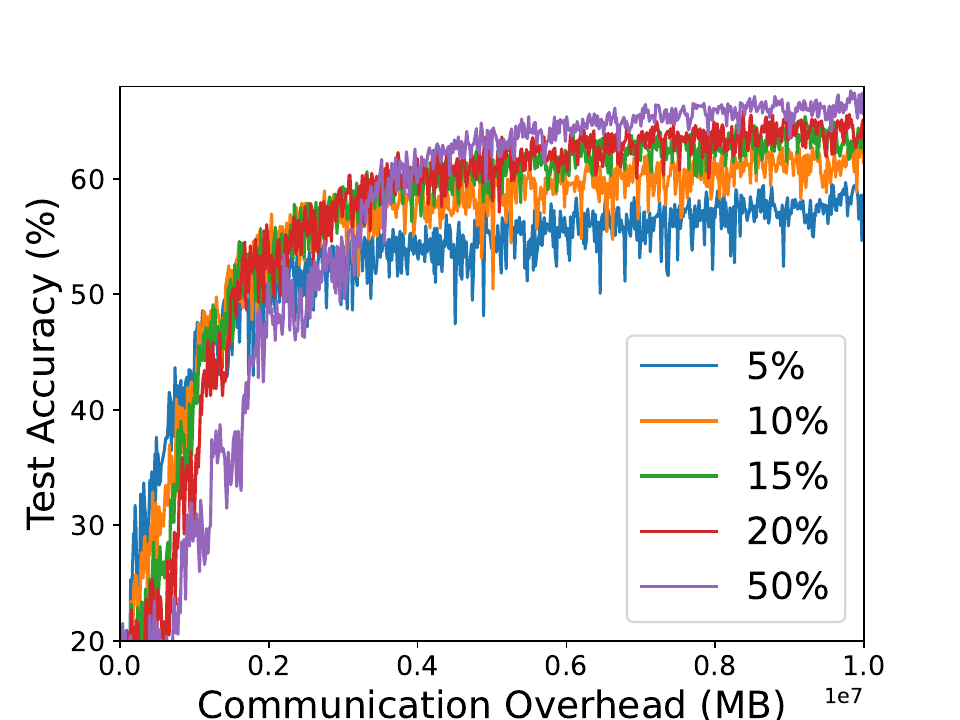}
		\caption{Impact of sampling proportion on test accuracy.}
		\label{fig:diff_p} 
	\end{center}
\end{figure}

{\bf Impacts of Sampling Proportion.}
In our approach, the sampling proportion plays an important role in determining 
 the global model performance. To evaluate the impacts of sampling proportion on the performance of the global model, we investigated the performance of our method with different sampling proportions, i.e.,  5\%, 10\%, 15\%, 20\%, and 50\%. 
 We conducted experiments on CIFAR-10 using the ResNet-18 model with $\beta=0.1$, whose results are shown in  Figure \ref{fig:diff_p}.
 We can find that a larger sampling proportion will lead to better test accuracy.
 Note that, as the sampling proportion increases, 
 the gain of test accuracy achieved by increasing the sampling proportion becomes smaller.


\begin{figure}[h]
  \begin{center} 
		\includegraphics[width=0.9\linewidth]{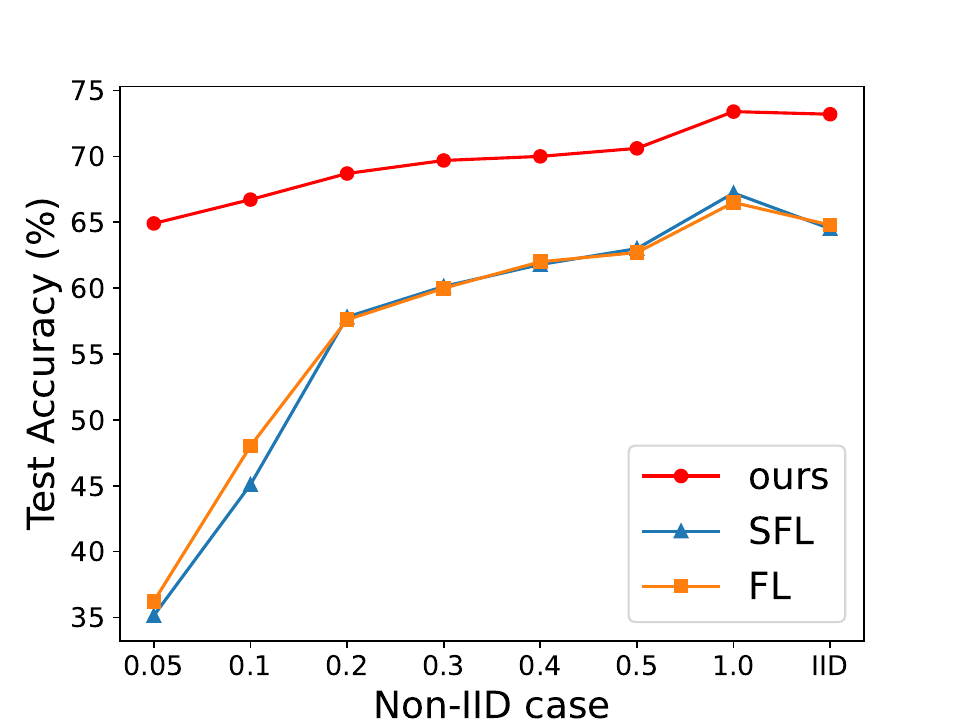}
 		\caption{Test accuracy under different non-IID distribution cases.}
		\label{fig:diff_b} 
	\end{center}
\end{figure}

{\bf Robustness in Non-IID  Scenarios.}
To verify the robustness of our method in non-IID scenarios, we conducted experiments on CIFAR-10 using the ResNet-18 model considering different non-IID settings, whose results are shown in  Figure \ref{fig:diff_b}. We can find that for a more non-IID scenario, 
SFL and FL suffer much more severe accuracy degradation than KoReA-SFL. 
Especially when dealing with the extreme non-IID case with $\beta=0.05$, 
we can observe a significant improvement in test accuracy achieved by KoReA-SFL compared with the other two counterparts. 
Note that for  KoReA-SFL, the test accuracy gap between the IID case and the non-IID case with $\beta=0.05$ is less than 9\%, while 
such gaps can hit around 30\%  for both SFL and FL. Obviously,  KoReA-SFL 
can mitigate the catastrophic forgetting problem in non-IID scenarios. 

\section{Conclusion}
\label{sec:conclusion}
This paper proposed a novel SFL framework named KoReA-SFL, aiming to address the notorious catastrophic forgetting problem when dealing with non-IID scenarios.
By continuously conducting local training on clients without being interrupted by the global model for aggregation, the training process of KoReA-SFL models approximately touches all the client data, thus mitigating the impact of non-IID distributions of client data. 
Meanwhile, by applying our proposed knowledge replay mechanism
and sampling strategy for assistant clients, 
the overall training process of KoReA-SFL can be accelerated while the underlying catastrophic forgetting symptom can be 
greatly suppressed. 
Besides the theoretical convergence analysis of KoReA-SFL,  we conducted comprehensive experiments to demonstrate the superiority of KoReA-SFL in improving the overall test accuracy.

\bibliographystyle{ACM-Reference-Format}
\bibliography{sample-base}

\end{document}